\pdfoutput=1

\documentclass[11pt]{article}

\usepackage[final]{acl}

\usepackage{times}
\usepackage{latexsym}
\usepackage{multirow}
\usepackage{booktabs}
\usepackage[T1]{fontenc}

\usepackage[utf8]{inputenc}
\usepackage{amsfonts}
\usepackage{textcomp}
\DeclareUnicodeCharacter{0394}{$\Delta$}
\usepackage{amsmath}
\usepackage{amssymb}
\usepackage{microtype}

\usepackage{inconsolata}

\usepackage{booktabs}  
\usepackage{amsmath}   
\usepackage[table]{xcolor}   
\usepackage{xfp}             

\usepackage{caption}   
\usepackage{multirow}  
\usepackage{graphicx}  
\usepackage{amssymb}   
\usepackage{makecell}  
\usepackage{ragged2e}  
\usepackage{tablefootnote}

\usepackage{enumitem}            
\usepackage{tikz}                
\usetikzlibrary{shapes.geometric, arrows.meta, positioning}

\definecolor{stepA}{RGB}{176,224,230} 
\definecolor{stepB}{RGB}{255,218,185} 
\definecolor{stepC}{RGB}{255,255,204} 
\definecolor{stepD}{RGB}{204,235,197} 
\definecolor{stepE}{RGB}{219,219,219} 

\definecolor{myteal}{HTML}{00897B}  

\usepackage[most]{tcolorbox}

\tcbset{
  parskip/.style={
    before={\par\pagebreak[0]\parindent=0pt},
    after={\parfillskip=0pt\par},
  },
}
\newtcolorbox{resultbox}[1][]{%
    colback=gray!30,
    colframe=black!3,
    notitle,
    sharp corners,
    borderline west={2pt}{0pt}{gray!80!black},
    enhanced,
    breakable,
    boxsep=0pt,
    left=4pt,right=2pt,top=2pt,bottom=2pt,
    }

\newtcolorbox{resultbox2}[1][]{%
  colback=gray!70,        
  colframe=white!30,      
  notitle,
  sharp corners,
  borderline west={2pt}{0pt}{gray!80},  
  enhanced,
  breakable,
  boxsep=0pt,
  left=4pt,right=2pt,top=2pt,bottom=2pt,
  #1
}

\newtcolorbox{resultbox3}[1][]{%
  colback=teal!70,        
  colframe=white!30,      
  notitle,
  sharp corners,
  borderline west={2pt}{0pt}{teal!80},  
  enhanced,
  breakable,
  boxsep=0pt,
  left=4pt,right=2pt,top=2pt,bottom=2pt,
  #1
}

\usepackage{graphicx}
\usepackage{blindtext}

\usepackage{geometry}
\usepackage{amssymb}
\usepackage{graphicx}
\usepackage{marvosym}
\usepackage{url}
\usepackage{fancyhdr}
\usepackage{hyperref}

\usepackage{enumitem}
\usepackage{booktabs}
\usepackage{algorithm}
\usepackage{algorithmic}
\usepackage{amsmath}
\usepackage{subcaption}
\usepackage{colortbl}
\usepackage{pifont}
\usepackage{colortbl}
\usepackage[table]{xcolor}

\newcommand{\xmark}{{\color{red}\textbf{X}}}
\newcommand{\cmark}{{\color{green}\textbf{Y}}}

\definecolor{class5}{RGB}{255, 204, 204} 
\definecolor{class4}{RGB}{255, 255, 204} 
\definecolor{class3}{RGB}{204, 255, 204} 
\definecolor{class2}{RGB}{204, 255, 255} 
\definecolor{class1}{RGB}{204, 204, 255} 
\definecolor{class0}{RGB}{255, 204, 255} 
\definecolor{NavyBlue}{RGB}{40, 160, 250} 

\newcommand{\sbs}[1]{%
    \pgfmathsetmacro{\normalized}{int((#1 - 0.4) / (1.0 - 0.4) * 100)}%
    \edef\tempa{\noexpand\cellcolor{NavyBlue!\normalized}}%
    \tempa#1%
}

\newcommand{\sbsalt}[1]{%
    \pgfmathsetmacro{\normalized}{int(min(100, max(0, #1 * 100)))}%
    \edef\tempa{\noexpand\cellcolor{NavyBlue!\normalized}}%
    \tempa#1%
}

\newcommand{\sbsal}[1]{%
    \pgfmathsetmacro{\normalized}{int(min(100, max(0, #1 * 100)))}%
    \edef\tempa{\noexpand\cellcolor{teal!\normalized}}%
    \tempa#1%
}

\newcommand{\sbsalty}[1]{%
    \pgfmathsetmacro{\normalized}{int(min(100, max(0, #1 * 100)))}%
    \edef\tempa{\noexpand\cellcolor{gray!\normalized}}%
    \tempa#1%
}

\newcommand{\updelta}[1]{\sbsa{#1}\,$\uparrow$}

\newcommand{\sbsa}[1]{%
  \pgfmathsetmacro{\normalized}{int(min(100, max(0, (#1/0.35)*100)))}%
  \edef\tempa{\noexpand\cellcolor{green!\normalized}}%
  \tempa #1%
}

\usepackage[normalem]{ulem}

\title{TigerCoder: A Novel Suite of LLMs for \\ Code Generation in Bangla}

\author{Nishat Raihan, Antonios Anastasopoulos, Marcos Zampieri \\ 
George Mason University \\ 
Fairfax, VA, USA \\
\texttt{mraihan2@gmu.edu}
}

\begin{document}
\maketitle
\begin{abstract}
Despite being the 5\textsuperscript{th} most spoken language, Bangla remains underrepresented in Large Language Models (LLMs), particularly for code generation. This primarily stems from the scarcity of high-quality data to pre-train and/or finetune such models. Hence, we introduce the first dedicated family of Code LLMs for Bangla (1B \& 9B). We offer three major contributions: (1) a comprehensive Bangla code instruction datasets for programming domain adaptation; (2) \texttt{MBPP-Bangla}, an evaluation benchmark for Bangla code generation; and (3) the \textbf{\texttt{TigerCoder}}-family of Code LLMs, achieving significant~11-18\% performance gains at \texttt{Pass@1} over existing multilingual and general-purpose Bangla LLMs. Our findings show that curated, high-quality datasets can overcome limitations of smaller models for low-resource languages. We open-source all resources\footnote{\url{https://github.com/mraihan-gmu/TigerCoder/}} to advance further Bangla LLM research.
\end{abstract}

\section{Introduction}

Recent advancements in LLMs have significantly improved code generation \cite{touvron2023llama, hui2024qwen2, team2025gemma}. State-of-the-art models often achieve over 90\% \textit{\texttt{Pass@1}} scores on popular coding benchmarks like HumanEval \cite{chen2021evaluating} and MBPP \cite{austin2021program} which led to their widespread adoption in domains such as software engineering \cite{pasquale2025challenges} and education \cite{raihan2025large}. This progress, however, disproportionately benefits high-resource languages \cite{joshi2020state, blasi2022systematic, ahuja2023mega, wang2023babel}. Specifically, Bangla, with over 242 million native speakers\footnote{\url{ethnologue.com/}} still lacks representation in code generation tasks. This deficiency originates from scarce datasets, limited resources, and absent benchmarks \cite{bhattacharjee2022banglabert, zehady2024bongllama}, leading to poor performance from existing general-purpose Bangla models compared to English counterparts \cite{bhattacharyya-etal-2023-vacaspati, uddin2023exploring}.

Current multilingual models like BLOOM \cite{scao2022bloom} and LLaMA-3 \cite{dubey2024llama} offer only basic Bangla support, exhibiting code generation performance as low as 20\% for Bangla compared to 65-75\% \textit{\texttt{Pass@1}} for English \cite{muennighoff2023crosslingual, raihan2024mhumaneval} due to minimal data allocation \cite{iyer2022optimizing, ahuja2024few}. To address this critical gap, we introduce \texttt{TigerCoder}, the first dedicated Bangla code-generation LLM (1B \& 9B parameters), finetuned on a large (\textit{300K instruction-code pairs}) Bangla Code Instruction dataset, tailored for the task and language. 

We address 2 research questions (RQs):

\begin{enumerate}[label=\textbf{\texttt{RQ\arabic*}}, leftmargin=*]
    \item To what extent do state-of-the-art Code LLMs preserve their code-generation quality when the natural language part of the prompt is written in \textit{Bangla} rather than English?

    \item Does a simple \textit{Bangla $\rightarrow$ English} machine-translation step applied to each coding prompt significantly boost generation quality compared with direct Bangla inference?

\end{enumerate}

\begin{table*}[!t]
\centering
\small
\scalebox{0.85}{
\begin{tabular}{l c c c c c c c}
\toprule
 & \textbf{Size} & \textbf{pt} & \textbf{corpora} & \textbf{ft}  & \textbf{ft-dataset} & \textbf{Paper?} & \textbf{Reprod.?} \\
\midrule
\href{https://huggingface.co/hishab/titulm-gemma-2-2b-v1.1}{titu-Gemma} & 2B & 4.4B & \xmark & \xmark & \xmark & \xmark & \xmark \\
\href{https://huggingface.co/hishab/titulm-llama-3.2-3b-v1.1}{titu-LLaMA} & 3B & 37B  & \xmark & \xmark & \xmark & \xmark & \xmark \\
\href{https://huggingface.co/BanglaLLM/BanglaLLama-3.2-3b-bangla-alpaca-orca-instruct-v0.0.1}{Bangla-LLaMA} & 3B & \cmark  & \xmark & 172K & Orca-trans. & \cmark & \xmark \\
\href{https://huggingface.co/KillerShoaib/gemma-2-9b-bangla-4bit}{G2B} & 9B & \xmark  & \xmark & 145K & Alpaca-trans. & \xmark & \xmark \\
\href{https://huggingface.co/BanglaLLM/bangla-llama-13b-instruct-v0.1}{Bangla-LLaMA} & 13B & \cmark & \xmark & 145K & Alpaca-trans. & \xmark & \xmark \\
\texttt{TigerLLM} & 9B & 10M & \texttt{Bangla-TB} & 100K & \texttt{Bangla-Inst} & \cmark & \cmark \\
\bottomrule
\end{tabular}
}
\caption{Comparative analysis of Bangla LLM initiatives and their methodological approaches. The pre-training (\textit{pt}) and finetuning (\textit{ft}) columns indicate corpus size in tokens and instruction count respectively.}
\label{table_models}
\end{table*}

\noindent With Bangla as our focus, we show how to develop small, open-source, high performance code generation models for low-resource languages. Our key contributions are the following:

\begin{enumerate}[label=\textbf{\texttt{C\arabic*}}, leftmargin=*, noitemsep]
    \item Development of three novel, high-quality Bangla code instruction datasets: \texttt{Bangla-Code-Instruct-SI} (100K, self-instruct), \texttt{Bangla-Code-Instruct-Syn} (100K, synthetic), and \texttt{Bangla-Code-Instruct-TE} (100K, machine-translated).
    \item Introduction of \texttt{MBPP-Bangla}, a benchmark with 974 expert-validated programming problems adopted from MBPP \cite{austin2021program}, designed specifically for evaluating Bangla code generation for 5 different programming languages (PLs) - \textit{Python, C++, JAVA, JavaScript \& Ruby}.
    \item The \textbf{\texttt{TigerCoder}} model family that outperforms existing models on Bangla programming tasks, achieving strong \textit{\texttt{Pass@1}} scores across all 5 PLs.
\end{enumerate}

\noindent The remainder of this paper is organized as follows: In Section \ref{sec:rw}, we discuss the current state of Bangla LLMs, then introduce \texttt{MBPP-Bangla} in Section \ref{sec:mbpp_bangla}; we address the two RQs in Section \ref{sec:rq1} \& \ref{sec:rq2}; then describe the curation of \texttt{Bangla-Code-Instruct} in Section \ref{sec:bcs} and \texttt{TigerCoder} in Section \ref{sec:tc}.

\section{Related Work} \label{sec:rw}

\paragraph{Bangla LLMs}
Table~\ref{table_models} catalogs the Bangla-centric LLMs released to date.  
Although these initiatives demonstrate encouraging progress, most models are distributed only as opaque checkpoints with limited details on data provenance, pre-processing, or training hyperparameters.  
This paucity of documentation hampers independent verification, reproducibility, and downstream adaptation—key pillars of scientific progress.

\paragraph{Bangla Benchmarks}
To assess Bangla NLP capability, researchers have introduced several task-specific benchmarks—e.g., BanglaRQA for reading comprehension \cite{ekram2022banglarqa}, BEnQA for open-domain question answering \cite{shafayat2024benqa}, and mHumanEval-Bangla for code generation \cite{raihan2024mhumaneval}.  
More recently, \citet{raihan2025tigerllm} released TigerLLM, a suite of general-purpose Bangla models trained on a carefully curated corpus, which set a stronger baseline across these datasets.

\paragraph{Code Generation in Bangla}
Despite the growth in Bangla NLU and NLG resources, code generation remains virtually unexplored.  
Prior work evaluates multilingual LLMs on Bangla prompts \cite{raihan2024mhumaneval} but no model has been purpose-built for this setting.  
TigerCoder addresses this gap by (i) constructing the first Bangla code-instruction corpus, (ii) training specialised Bangla Code LLM checkpoints, and (iii) establishing a rigorous evaluation pipeline that disentangles programming competence from Bangla comprehension.  
Our work therefore extends the TigerLLM lineage into the software-engineering domain and provides a reproducible foundation for future research on low-resource code generation.

\section{The \texttt{MBPP}-Bangla Benchmark}
\label{sec:mbpp_bangla}

\paragraph{Structure}
To address the absence of standardized evaluation tools for Bangla–code generation, we introduce \textbf{\texttt{MBPP}-Bangla}.  Derived from the Mostly Basic Python Programs (MBPP) dataset \cite{austin2021program}, \texttt{MBPP}-Bangla comprises \textbf{974} programming problems, each presented in Bangla and paired with \emph{canonical reference solutions} in \textbf{five languages}: Python, Java, JavaScript, Ruby, and~C++.  
In addition, every problem is assigned to one of five topical classes (\textit{String, Math, Data-structures, Algorithms, File-I/O}), enabling controlled, category-wise evaluation.

\paragraph{Curation Process} The complete curation phase goes through five key steps - 
\begin{enumerate}[label=\textbf{\texttt{Step\arabic*}}, leftmargin=*, noitemsep, nolistsep]
  \item We first extract all 974 basic–intermediate problems from the Mostly Basic Python Programs (MBPP) corpus, ensuring coverage of five topical classes.

  \item Two native Bangla speakers, each with certified English proficiency (TOEFL \textit{Internet‐based\,—\,score $>$\,100}), translate every English prompt into Bangla without consulting one another.

  \item For each task we curate or port five reference solutions—Python, Java, JavaScript, Ruby, and C++—so that downstream systems can be evaluated language-by-language.

  \item A third reviewer, fluent in Bangla and all five programming languages, manually checks every item for linguistic fidelity, technical correctness, and cross-language consistency.

  \item The validated records are released with a task ID, Bangla prompt, five reference codes, original MBPP test cases, and a topic label, forming the \texttt{MBPP}-Bangla benchmark.
\end{enumerate}

\noindent Table~\ref{tab:mbpp_bangla_params} summarizes the key parameters of the curation process. The full curation protocol, topical taxonomy, and multilingual solution-conversion pipeline are documented in Appendix~\ref{appendix:mbpp_bangla}.%
\footnotetext[1]{\label{fn:toefl}\url{https://www.ets.org/toefl.html}}

\begin{table}[!t]
\centering
\small
\scalebox{0.95}{
\begingroup
\setlength{\tabcolsep}{6pt}
\begin{tabular}{%
    >{\columncolor{gray!50}}p{0.32\linewidth}%
    >{\columncolor{gray!30}}p{0.60\linewidth}}
\toprule
\rowcolor{gray!70}
\multicolumn{1}{c}{\bf Parameter} & \multicolumn{1}{c}{\bf Specification} \\ \midrule
Dataset Size            & 974 problems \\
Source                  & Mostly Basic Python Programs (MBPP) \cite{austin2021program} \\
Programming Languages   & Python, Java, JavaScript, Ruby, C++ \\
Task Focus              & Basic–Intermediate coding problems (five topical classes) \\
Translation Method      & Human translation (independent, manual) \\
Translators             & 2 native Bangla speakers (worked independently) \\
Translator Proficiency  & Bangla (native), English (TOEFL$^{\ref{fn:toefl}}$ $>$100) \\
Verification Method     & Human expert verification (manual) \\
Verifier                & 1 native Bangla speaker \& polyglot programmer \\
Verifier Expertise      & Proficient in all five languages above \\
Verification Scope      & Linguistic fidelity, Technical correctness, Cross‐lang.\ consistency \\
Output Components       & Task ID, Bangla prompt, 5× reference codes, test cases, topic class \\
\bottomrule
\end{tabular}
\endgroup}
\caption{Curation parameters for \texttt{MBPP}-Bangla.  Both columns use a light‐gray background, with a slightly darker shade for the header row.}
\label{tab:mbpp_bangla_params}
\end{table}


\section{RQ1 - Evaluating LLMs on Bangla Code Generation Benchmarks}
\label{sec:rq1}

Several works on LLM evaluation have already revealed that these models often perform better when prompted in English \cite{rohera2025better, li2024quantifying, schut2025multilingual}. None of them have investigated this phenomenon for the task of \textit{Code Generation}. In this section, we attempt to shed some light on it.

\paragraph{Benchmarks}
We consider both \texttt{MBPP-Bangla} \& \texttt{mHumanEval-Bangla} for evaluation purposes for their unique design. Compared with \texttt{mHumanEval-ben}, \texttt{MBPP}-Bangla offers complementary benefits. Its larger size yields more statistically robust estimates, and its conversational Bangla prompts—spanning five programming languages—stress a model’s ability to \emph{both} comprehend varied natural language specifications and generate syntactically valid code in multiple target languages. Employing the two benchmarks jointly therefore affords a comprehensive assessment of Bangla code-generation capability.

\paragraph{LLMs}
For the evaluation phase, we choose all the released Bangla LLMs - listed in Table \ref{table_models}. We also consider the multilingual open-source models, including LLaMA 3.2 \cite{dubey2024llama}, Gemma 3 \cite{gemma3technicalreport}, Phi 4 \cite{phi4}, and Pangea \cite{yue2024pangea}. In addition, we evaluate proprietary LLMs of similar size, like GPT-4o-mini \cite{openai2024gpt4o}, Gemini-2.5-Flash \cite{google2024gemini2.5}, and GPT-3.5 \cite{wang2023comprehensive}.

\begin{table*}[!t]
\centering
\small
\scalebox{0.8}{
\begin{tabular}{l*{6}{c}|*{6}{c}}

\toprule
\multirow{3}{*}{\textbf{Model}} &
\multicolumn{6}{c}{\textbf{mHumanEval}} &
\multicolumn{6}{c}{\textbf{MBPP}} \\
\cmidrule(lr){2-7}\cmidrule(lr){8-13}
& \multicolumn{3}{c}{\textit{English}} & \multicolumn{3}{c}{\textit{Bangla}} &
  \multicolumn{3}{c}{\textit{English}} & \multicolumn{3}{c}{\textit{Bangla}} \\
\cmidrule(lr){2-4}\cmidrule(lr){5-7}\cmidrule(lr){8-10}\cmidrule(lr){11-13}
& P@1 & P@10 & P@100 & P@1 & P@10 & P@100 &
  P@1 & P@10 & P@100 & P@1 & P@10 & P@100 \\
\midrule
GPT-3.5                      & \sbsalt{0.79} & \sbsalt{0.81} & \sbsalt{0.84} & \sbsalt{0.56} & \sbsalt{0.56} & \sbsalt{0.59} & \sbsalt{0.81} & \sbsalt{0.83} & \sbsalt{0.89} & \sbsalt{0.60} & \sbsalt{0.62} & \sbsalt{0.62} \\
Gemini-Flash 2.5             & \sbsalt{0.82} & \sbsalt{0.85} & \sbsalt{0.89} & \sbsalt{0.58} & \sbsalt{0.61} & \sbsalt{0.62} & \sbsalt{0.82} & \sbsalt{0.85} & \sbsalt{0.91} & \sbsalt{0.62} & \sbsalt{0.62} & \sbsalt{0.70} \\
GPT-4o-mini                  & \sbsalt{0.74} & \sbsalt{0.79} & \sbsalt{0.84} & \sbsalt{0.56} & \sbsalt{0.56} & \sbsalt{0.58} & \sbsalt{0.77} & \sbsalt{0.84} & \sbsalt{0.87} & \sbsalt{0.51} & \sbsalt{0.53} & \sbsalt{0.54} \\
\midrule
LLaMA-3.2 (11B)              & \sbsalt{0.73} & \sbsalt{0.76} & \sbsalt{0.77} & \sbsalt{0.15} & \sbsalt{0.15} & \sbsalt{0.20} & \sbsalt{0.78} & \sbsalt{0.81} & \sbsalt{0.86} & \sbsalt{0.22} & \sbsalt{0.22} & \sbsalt{0.30} \\
Gemma-3 (27B)                & \sbsalt{0.69} & \sbsalt{0.71} & \sbsalt{0.78} & \sbsalt{0.64} & \sbsalt{0.65} & \sbsalt{0.69} & \sbsalt{0.71} & \sbsalt{0.78} & \sbsalt{0.83} & \sbsalt{0.69} & \sbsalt{0.70} & \sbsalt{0.70} \\
Pangea (7B)                  & \sbsalt{0.61} & \sbsalt{0.63} & \sbsalt{0.63} & \sbsalt{0.10} & \sbsalt{0.15} & \sbsalt{0.20} & \sbsalt{0.64} & \sbsalt{0.65} & \sbsalt{0.65} & \sbsalt{0.09} & \sbsalt{0.12} & \sbsalt{0.17} \\
Phi-4 (7B)                   & \sbsalt{0.79} & \sbsalt{0.81} & \sbsalt{0.86} & \sbsalt{0.10} & \sbsalt{0.17} & \sbsalt{0.25} & \sbsalt{0.82} & \sbsalt{0.84} & \sbsalt{0.89} & \sbsalt{0.09} & \sbsalt{0.15} & \sbsalt{0.20} \\
\midrule
Titu-LLM (2B)                & \sbsalt{0.20} & \sbsalt{0.20} & \sbsalt{0.23} & \sbsalt{0.02} & \sbsalt{0.02} & \sbsalt{0.02} & \sbsalt{0.21} & \sbsalt{0.23} & \sbsalt{0.23} & \sbsalt{0.05} & \sbsalt{0.05} & \sbsalt{0.05} \\
Bong-LLaMA (3B)              & \sbsalt{0.31} & \sbsalt{0.32} & \sbsalt{0.34} & \sbsalt{0.02} & \sbsalt{0.02} & \sbsalt{0.02} & \sbsalt{0.04} & \sbsalt{0.04} & \sbsalt{0.04} & \sbsalt{0.07} & \sbsalt{0.09} & \sbsalt{0.11} \\
Bangla-LLaMA (3B)            & \sbsalt{0.44} & \sbsalt{0.48} & \sbsalt{0.50} & \sbsalt{0.10} & \sbsalt{0.19} & \sbsalt{0.21} & \sbsalt{0.41} & \sbsalt{0.49} & \sbsalt{0.55} & \sbsalt{0.13} & \sbsalt{0.15} & \sbsalt{0.15} \\
Bangla-Gemma (9B)            & \sbsalt{0.47} & \sbsalt{0.49} & \sbsalt{0.49} & \sbsalt{0.10} & \sbsalt{0.15} & \sbsalt{0.16} & \sbsalt{0.45} & \sbsalt{0.49} & \sbsalt{0.50} & \sbsalt{0.08} & \sbsalt{0.19} & \sbsalt{0.21} \\
\midrule
\textit{TigerLLM} (1B)       & \sbsalt{0.64} & \sbsalt{0.66} & \sbsalt{0.66} & \sbsalt{0.61} & \sbsalt{0.64} & \sbsalt{0.70} & \sbsalt{0.68} & \sbsalt{0.69} & \sbsalt{0.69} & \sbsalt{0.65} & \sbsalt{0.68} & \sbsalt{0.71} \\
\textit{TigerLLM} (9B)       & \sbsalt{0.69} & \sbsalt{0.69} & \sbsalt{0.71} & \sbsalt{0.63} & \sbsalt{0.69} & \sbsalt{0.72} & \sbsalt{0.71} & \sbsalt{0.73} & \sbsalt{0.75} & \sbsalt{0.61} & \sbsalt{0.68} & \sbsalt{0.73} \\
\bottomrule
\end{tabular}}
\caption{Pass@\{1,10,100\} comparison on English and Bangla variants of \texttt{mHumanEval} and \texttt{MBPP}. It shows the consistent subpar performance of most models (except for \texttt{TigerLLM}) in Bangla, compared to English in code generation. (The darker the cell color, the better the performance.)}
\label{tab:code_results_full}
\end{table*}

\paragraph{Metric}
We use \textit{Pass@K} as our evaluation metric. For a task with $n$ generated programs, $m$ of which pass all tests, \textit{Pass@K} is defined as -

\[
\mathrm{Pass@}K \;=\; 1 - \frac{\binom{\,n-m\,}{K}}{\binom{n}{K}}, 
\qquad 1 \le K \le n.
\]
It is the probability that at least one of $K$ \emph{without-replacement} draws is correct.

We adopt 3 variations of \textit{Pass@K}:
\paragraph{$K\!\in\!\{1,10,100\}$.}
\begin{itemize}[nosep,leftmargin=1.4em]
\item $K=1$: measures single-shot quality (default user experience).
\item $K=10$: reflects a small, practical shortlist for manual inspection.
\item $K=100$: estimates the model’s upper potential under heavy sampling.
\end{itemize}

\paragraph{Observation}
The results reveal a consistent performance gap, with most models favoring English over Bangla. While proprietary models excel in English, they are outperformed by \texttt{TigerLLM} in Bangla. This performance drop in Bangla is even more pronounced among the open-source models. Gemma 3 is the exception in this group, showing only a moderate decline.

The Bangla-specific models underperform significantly in English and perform even worse in Bangla. This leaves \texttt{TigerLLM} as the only model with similar performance in Bangla compared to English.

\paragraph{Analysis}
Most proprietary and large open-source models are trained on corpora dominated by English, so they naturally excel on the English subsets of both benchmarks \cite{commoncrawl}. Since Bangla appears only sparsely—and often without paired code examples—these models struggle to map Bangla instructions to valid programs, which leads to the sharp drops we see across \texttt{mHumanEval} and \texttt{MBPP}. Gemma 3 (27 B) shows a smaller gap because its greater capacity and multilingual pre-training let it transfer knowledge more effectively, but the English-first bias remains clear. In contrast, \texttt{TigerLLM} is fine-tuned on a purpose-built Bangla–instruction corpus, so it retains almost all of its English-side skill while markedly boosting Bangla performance—even though it is one to two orders of magnitude smaller than the proprietary systems.

Open-source Bangla-specific models (e.g., Bong-LLaMA, Bangla-Gemma) fare poorly because their training data are small, noisy, or text-only; without consistent, executable code pairs, they cannot learn the rigorous syntax and logical structure that Pass@K rewards. \texttt{TigerLLM’s} curated data close this gap: its 1 B variant rivals much larger multilingual models in Bangla, and the 9 B version narrows the English gap to within a few points while establishing a clear Bangla lead. These results underscore a practical lesson for low-resource languages—high-quality, domain-specific data outweigh both model size and generic multilingual pre-training when the task demands precise code generation.

\begin{table*}[!t]
\centering
\small
\scalebox{0.8}{
\begin{tabular}{l*{6}{c}|*{6}{c}}
\toprule
\multirow{3}{*}{\textbf{Model}} &
\multicolumn{6}{c}{\textbf{mHumanEval}} &
\multicolumn{6}{c}{\textbf{MBPP}} \\
\cmidrule(lr){2-7}\cmidrule(lr){8-13}
& \multicolumn{3}{c}{\textit{\textbf{Bangla $\rightarrow$ English}-MT}} & \multicolumn{3}{c}{Bangla} &
\multicolumn{3}{c}{\textit{\textbf{Bangla $\rightarrow$ English}-MT}} & \multicolumn{3}{c}{Bangla} \\
\cmidrule(lr){2-4}\cmidrule(lr){5-7}\cmidrule(lr){8-10}\cmidrule(lr){11-13}
& P@1 & P@10 & P@100 & P@1 & P@10 & P@100 &
P@1 & P@10 & P@100 & P@1 & P@10 & P@100 \\
\midrule
GPT-3.5 & \sbsal{0.48} & \sbsal{0.44} & \sbsal{0.53} & \sbsal{0.56} & \sbsal{0.56} & \sbsal{0.59} & \sbsal{0.55} & \sbsal{0.50} & \sbsal{0.67} & \sbsal{0.60} & \sbsal{0.62} & \sbsal{0.62} \\
Gemini-Flash 2.5 & \sbsal{0.51} & \sbsal{0.63} & \sbsal{0.53} & \sbsal{0.58} & \sbsal{0.61} & \sbsal{0.62} & \sbsal{0.58} & \sbsal{0.66} & \sbsal{0.61} & \sbsal{0.62} & \sbsal{0.62} & \sbsal{0.70} \\
GPT-4o-mini & \sbsal{0.45} & \sbsal{0.59} & \sbsal{0.49} & \sbsal{0.56} & \sbsal{0.56} & \sbsal{0.58} & \sbsal{0.43} & \sbsal{0.46} & \sbsal{0.42} & \sbsal{0.51} & \sbsal{0.53} & \sbsal{0.54} \\
\midrule
LLaMA-3.2 (11B) & \sbsal{0.10} & \sbsal{0.11} & \sbsal{0.11} & \sbsal{0.15} & \sbsal{0.15} & \sbsal{0.20} & \sbsal{0.19} & \sbsal{0.12} & \sbsal{0.18} & \sbsal{0.22} & \sbsal{0.22} & \sbsal{0.30} \\
Gemma-3 (27B) & \sbsal{0.59} & \sbsal{0.56} & \sbsal{0.62} & \sbsal{0.64} & \sbsal{0.65} & \sbsal{0.69} & \sbsal{0.64} & \sbsal{0.60} & \sbsal{0.73} & \sbsal{0.69} & \sbsal{0.70} & \sbsal{0.70} \\
Pangea (7B) & \sbsal{0.00} & \sbsal{0.03} & \sbsal{0.14} & \sbsal{0.10} & \sbsal{0.15} & \sbsal{0.20} & \sbsal{0.03} & \sbsal{0.07} & \sbsal{0.05} & \sbsal{0.09} & \sbsal{0.12} & \sbsal{0.17} \\
Phi-4 (7B) & \sbsal{0.04} & \sbsal{0.06} & \sbsal{0.22} & \sbsal{0.10} & \sbsal{0.17} & \sbsal{0.25} & \sbsal{0.00} & \sbsal{0.03} & \sbsal{0.13} & \sbsal{0.09} & \sbsal{0.15} & \sbsal{0.20} \\
\midrule
Titu-LLM (2B) & \sbsal{0.00} & \sbsal{0.00} & \sbsal{0.00} & \sbsal{0.02} & \sbsal{0.02} & \sbsal{0.02} & \sbsal{0.00} & \sbsal{0.01} & \sbsal{0.00} & \sbsal{0.05} & \sbsal{0.05} & \sbsal{0.05} \\
Bong-LLaMA (3B) & \sbsal{0.00} & \sbsal{0.00} & \sbsal{0.00} & \sbsal{0.02} & \sbsal{0.02} & \sbsal{0.02} & \sbsal{0.11} & \sbsal{0.12} & \sbsal{0.02} & \sbsal{0.07} & \sbsal{0.09} & \sbsal{0.11} \\
Bangla-LLaMA (3B) & \sbsal{0.01} & \sbsal{0.07} & \sbsal{0.12} & \sbsal{0.10} & \sbsal{0.19} & \sbsal{0.21} & \sbsal{0.04} & \sbsal{0.09} & \sbsal{0.12} & \sbsal{0.13} & \sbsal{0.15} & \sbsal{0.15} \\
Bangla-Gemma (9B) & \sbsal{0.04} & \sbsal{0.18} & \sbsal{0.04} & \sbsal{0.10} & \sbsal{0.15} & \sbsal{0.16} & \sbsal{0.03} & \sbsal{0.10} & \sbsal{0.15} & \sbsal{0.08} & \sbsal{0.19} & \sbsal{0.21} \\
\midrule
\textit{TigerLLM} (1B) & \sbsal{0.52} & \sbsal{0.58} & \sbsal{0.61} & \sbsal{0.61} & \sbsal{0.64} & \sbsal{0.70} & \sbsal{0.69} & \sbsal{0.57} & \sbsal{0.61} & \sbsal{0.65} & \sbsal{0.68} & \sbsal{0.71} \\
\textit{TigerLLM} (9B) & \sbsal{0.51} & \sbsal{0.74} & \sbsal{0.60} & \sbsal{0.63} & \sbsal{0.69} & \sbsal{0.72} & \sbsal{0.58} & \sbsal{0.63} & \sbsal{0.75} & \sbsal{0.61} & \sbsal{0.68} & \sbsal{0.73} \\
\bottomrule
\end{tabular}}
\caption{Pass@\{1,10,100\} comparison on Bangla and \texttt{Bangla$\rightarrow$English-MT} variants of \texttt{mHumanEval} and \texttt{MBPP}. We observe a similar or worse set of results across the board with MT prompts compared to Bangla.}
\label{tab:code_results_full2}
\end{table*}



\section{RQ2 - Does a simple \textit{Bangla $\rightarrow$ English} machine-translation help?} \label{sec:rq2}

As some of the recent works \cite{toukmaji2025prompt} have shown, LLMs often perform better when the prompts are simply translated to English before feeding them to the model. Again, this is also a scenario that is not explored in the domain of \textit{Code Generation}. In this section, we attempt to provide some careful insights into our second RQ, investigating whether a simple machine translation for the prompts will solve the issue of the performance drop, as shown in Table \ref{tab:code_results_full}. We adopt a similar experimental setup with the same benchmarks, models, and metrics from the previous experiment, detailed in Section \ref{sec:rq1}.

\paragraph{Prompt Translation}
We curate two new benchmarks from our two existing ones—\texttt{mHumanEval-ben} \& \texttt{MBPP-Bangla}—which are human-generated. We use NLLB \cite{costa2022nollb} for the translation, as this is the SOTA model for \textit{Bangla $\rightarrow$ English} machine-translation. Thus, we compile mHumanEval-\texttt{MT} \& MBPP-\texttt{MT}. We then carry out the similar set of experiments as Section \ref{sec:rq1}, but this time contrast between Bangla and Bangla-\texttt{MT} (Bangla$\rightarrow$English) variants of the benchmarks. 

\begin{figure}[!t]
    \centering
    \includegraphics[width=0.4\textwidth]{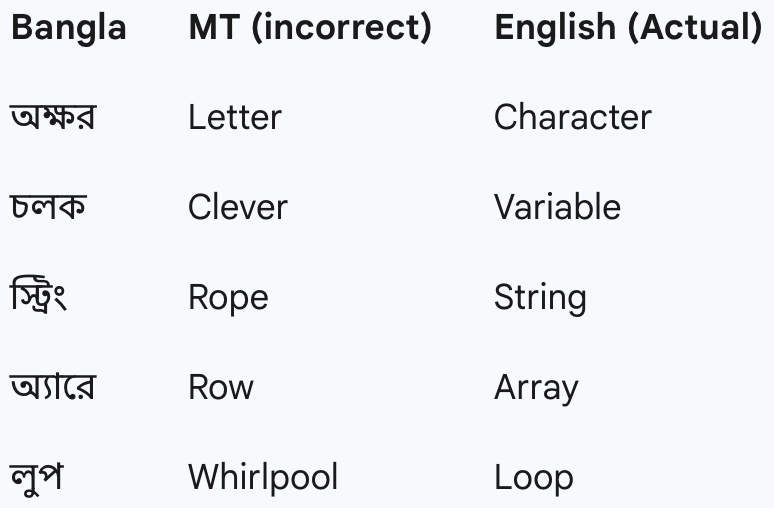}
    \caption{Incorrect keywords generated by machine translation.}
    \label{fig:keywords}
\end{figure}

\paragraph{Observation}
Unlike the previous set of experiments, we do not notice any significant performance decline; rather, the results are mostly similar or a bit worse with English-MT variants to Bangla. While the proprietary models do better in general, their performance does not match the original English benchmark (Table \ref{tab:code_results_full}), the same with all the other families of models. Hence, we can empirically compare the LLMs' performance over different variants of the same prompts as follows:

\begin{resultbox}
    \begin{center}
        Performance when prompted in \textbf{English} >
        Performance when prompted in \textbf{Bangla} >
        Performance when prompted in \textbf{English-MT}
    \end{center}
\end{resultbox}

\begin{table*}[!t]
\centering
\scalebox{0.87}{%
  \begin{minipage}{0.98\linewidth} 
  \centering
  \small 
  \begin{tabular}{@{} >{\RaggedRight}p{0.25\linewidth} >{\RaggedRight}p{0.24\linewidth} >{\RaggedRight}p{0.24\linewidth} >{\RaggedRight}p{0.24\linewidth} @{}}
  \toprule
  \textbf{\texttt{Bangla-Code-Instruct}} & \textbf{-SI} \newline\textit{(Self-Instruct)} & \textbf{-Syn} \newline\textit{(Synthetic)} & \textbf{-TE} \newline\textit{(Translated)} \\
  \midrule
  Size & 100,000 & 100,000 & 100,000 \\
  Method & Self-Instruction & Synthetic Generation & MT + Filtering \\
  Seed/Source & 5000 \textit{(Expert)} & Set of Topics & \texttt{Evol-Instruct} \\
  Teacher Model(s) & GPT-4o & GPT-4o \& Claude-3.5 & --- \\
  MT Model(s) & --- & --- & NLLB-200 \\
  Code Validation & Syntax + Execution Check & Syntax + Execution Check & Retained Source Code \\
  Filtering Metric(s) & Cosine Similarities & BERTScore & BERTScore + Comet QE\\
  Prompt Origin & Semi-Natural & Synthetic & Translated \\
  Code Origin & Synthetic & Synthetic & Natural (Source) \\
  \bottomrule
  \end{tabular}
  \caption{Comparing details of the three subsets of \texttt{Bangla-Code-Instruct}; \textbf{-SI}, \textbf{-Syn} \& \textbf{-TE}.}
  \label{tab:all_ins}
  \end{minipage}%
} 
\end{table*}

\paragraph{Analysis}
As we investigate the poor results with MT prompts, we notice a very specific trend leading to the generation of unexpected performance. As the coding prompts are translated into English prompts, several code-specific keywords are often translated into words that do not retain the same meaning. A few common examples are shown in Figure \ref{fig:keywords}, which fails to describe to task and misleads the models to generate poor results. 



\section{\texttt{Bangla-Code-Instruct}} \label{sec:bcs}

As is evident from the experiments of Sections \ref{sec:rq1} \& \ref{sec:rq2}, recent LLMs perform poorly with Bangla coding tasks, and the issue can not be solved with MT. Hence, we curate three instruction-tuning datasets in an attempt to finetune LLMs for this particular task and language. The datasets are tailored for Bangla: \texttt{Bangla Code Instruct-SI}, \texttt{-Syn}, and \texttt{-TE}. These datasets are specifically designed to capture diverse aspects of code generation and instruction understanding, ensuring a robust training foundation (see Table \ref{tab:all_ins}).

\subsection{\texttt{Bangla-Code-Instruct-SI}}
This dataset consists of 100,000 instruction-code pairs generated via self-instruction \cite{wang2023self}. The process begins with 5000 seed prompts manually authored in Bangla by programming experts. These are then used to generate a larger set of `semi-natural' instructional prompts (human-seeded and LLM-evolved using \texttt{GPT-4o}), also in Bangla. The corresponding Python code for each instruction is generated by \texttt{GPT-4o} and also \textit{execution-validated}, signifying that it passed both a syntax check (using \texttt{ast.parse}\footnote{\url{docs.python.org/3/library/ast.html}}) and successful execution in a controlled environment (\textit{Python 3.13.0}
\textit{, 10s timeout, 16GB memory}). The curation process and parameters are detailed in Appendix~\ref{appendix:si}.

\subsection{\texttt{Bangla-Instruct-Syn}}
This subset provides 100,000 synthetic Bangla instruction-Python code pairs generated by \texttt{GPT-4o} \cite{openai2023gpt4} and Claude 3.5-Sonnet \cite{anthropic2023claude}. To ensure instructional diversity, new instructions are compared against existing ones; a BERTScore \cite{zhang2020bertscore} of $\ge 0.7$ against any existing instruction results in the new pair being discarded. The LLMs are prompted in Bangla to produce these Bangla instructions and corresponding code for diverse tasks. For this synthetic set, code is validated for syntax (\texttt{ast.parse}) and execution (similar to -SI), aiming to broaden task diversity. This set complements the human-seeded data, though the naturalness of LLM-generated Bangla may differ from expert-authored versions. Appendix~\ref{appendix:syn} contains further generation and filtering details.

\begin{figure*}[!t]
    \centering
    \includegraphics[width=0.8\textwidth]{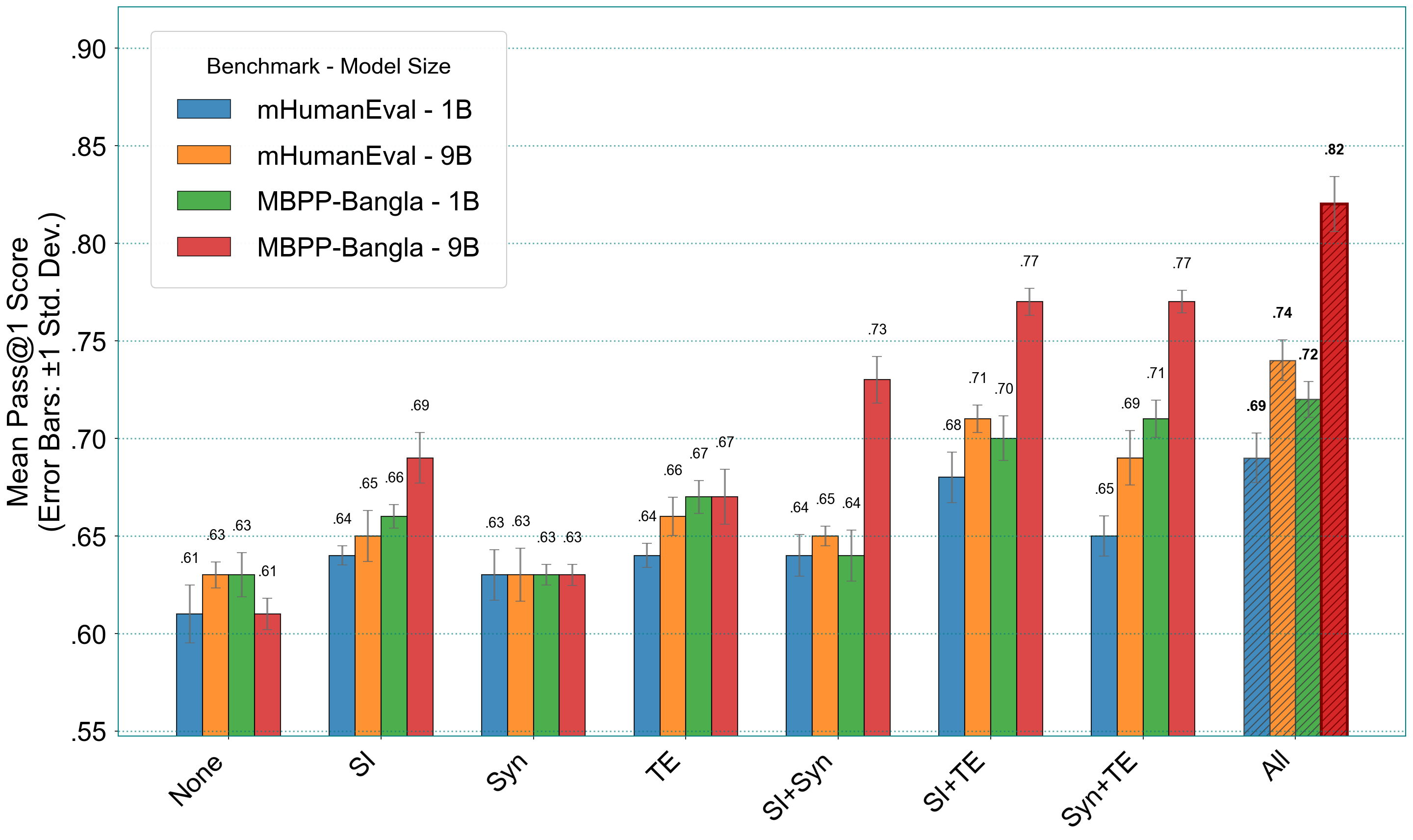}
    \caption{Performance (Pass@1) comparison for different combinations of the \texttt{SI}, \texttt{Syn}, and \texttt{TE} instruction datasets across model sizes (1B vs 9B).}
    \label{fig:mhuman_ablation}
\end{figure*}

\subsection{\texttt{Bangla-Instruct-TE}}
The final subset contains 100,000 prompt-code pairs by translating English instructions from Evol-Instruct \cite{xu2023wizardlm} using multiple MT models and selecting the best translation based on CometKiwi-22 QE \cite{rei2020cometkiwi} (> 0.85) and BERTScore F1 (> 0.95). The original code is retained. The process is specified in Appendix~\ref{appendix:te}.

The combined dataset (Table \ref{tab:all_ins}) of 300,000 examples provides \texttt{TigerCoder} with diverse training signals from human-seeded (\texttt{SI}), purely synthetic (\texttt{Syn}), and translation-based (\texttt{TE}) sources. 

\section{\texttt{TigerCoder}} \label{sec:tc}
\label{sec:tigercoder}

We choose TigerLLM \cite{raihan2025tigerllm} as our base model, as the results from Table \ref{tab:code_results_full} \& \ref{tab:code_results_full2} strongly indicate its efficacy in Bangla code generation. We finetune it using \texttt{Bangla-Code-Instruct} to build \texttt{TigerCoder} family that represents the first dedicated family of Large Language Models specifically optimized for Bangla code generation tasks.

\paragraph{Experimental Setup}
We conduct finetuning on a single NVIDIA A100 (40GB) through Google Colab\footnote{\url{colab.research.google.com}}, supported by 80GB RAM and 256GB storage. The process completes in approximately 96 hours, proving sufficient for model adaptation and task-specific optimization with minimal computational overhead.

\paragraph{Fine-tuning}
Given three instruction datasets with distinct curation methods, we fine-tune models on each individually and in all possible combinations for thorough analysis. Figure~\ref{fig:mhuman_ablation} shows strong base model performance (especially 1B). Fine-tuning on single datasets generally improves results, with \texttt{SI} and \texttt{TE} proving more effective than \texttt{Syn} alone. Dataset combinations exhibit clear synergistic effects: the \texttt{SI} + \texttt{TE} pairing significantly boosts performance, achieving top scores for the 9B model on \texttt{mHumanEval} (tying with all three datasets). Using all datasets (\texttt{SI} + \texttt{Syn} + \texttt{TE}) yields the best overall results across both benchmarks and model sizes, pushing the 9B model to 0.82 Pass@1 on \texttt{MBPP-Bangla}. The larger 9B model benefits more from comprehensive fine-tuning, particularly with combined datasets, compared to the 1B model. Performance on \texttt{MBPP-Bangla} is generally higher than on \texttt{mHumanEval} across most configurations.

\begin{table}[!t]
\centering
\small
\scalebox{0.9}{
\begingroup
\setlength{\tabcolsep}{6pt}
\begin{tabular}{%
    >{\columncolor{gray!50}}l
    >{\columncolor{gray!30}}c
    >{\columncolor{gray!30}}c}
\toprule
\rowcolor{gray!70}
\textbf{Hyperparameter} & \textbf{1B} & \textbf{9B} \\ \midrule
Max Sequence Length           & 2048            & 2048 \\
Batch Size (Train / Eval)      & 16              & 32   \\
Gradient Accum. Steps          & 4               & 8    \\
Number of Epochs               & 3               & 3    \\
Learning Rate                  & $1\times10^{-5}$ & $1\times10^{-6}$ \\
Weight Decay                   & 0.02            & 0.04 \\
Warm-up Steps                  & 10\%            & 15\% \\
Optimizer                      & AdamW   & AdamW \\
LR Scheduler                   & Cosine          & Cosine \\
Precision                      & BF16            & BF16 \\
Evaluation Strategy            & Steps           & Steps \\
Evaluation Steps               & 50              & 250  \\
Save Strategy                  & Steps           & Steps \\
Save Steps                     & Varies          & Varies \\
Seed                           & 42              & 42   \\
\bottomrule
\end{tabular}
\endgroup}
\caption{Empirically selected hyperparameters for fine-tuning the \texttt{TigerCoder} model family.}
\label{tab:hyperparameters_combined}
\end{table}

\begin{table*}[!t]
\centering
\small
\scalebox{0.85}{
\begin{tabular}{l*{6}{c}|*{6}{c}}
\toprule
\multirow{3}{*}{\textbf{}} &
\multicolumn{6}{c}{\textbf{mHumanEval} \textit{Bangla}} &
\multicolumn{6}{c}{\textbf{MBPP} \textit{Bangla}} \\
\cmidrule(lr){2-7}\cmidrule(lr){8-13}
& \multicolumn{2}{c}{P@1} & \multicolumn{2}{c}{P@10} & \multicolumn{2}{c}{P@100} &
  \multicolumn{2}{c}{P@1} & \multicolumn{2}{c}{P@10} & \multicolumn{2}{c}{P@100} \\
\cmidrule(lr){2-3}\cmidrule(lr){4-5}\cmidrule(lr){6-7}
\cmidrule(lr){8-9}\cmidrule(lr){10-11}\cmidrule(lr){12-13}
& Score & $\Delta$ & Score & $\Delta$ & Score & $\Delta$ &
  Score & $\Delta$ & Score & $\Delta$ & Score & $\Delta$ \\
\midrule
\texttt{\textbf{TigerCoder}} (1B) & 0.69 & \updelta{0.05} & 0.73 & \updelta{0.04} & 0.77 & \updelta{0.05} &
                         0.74 & \updelta{0.05} & 0.74 & \updelta{0.04} & 0.81 & \updelta{0.08} \\
\texttt{\textbf{TigerCoder}} (9B) & 0.75 & \updelta{0.11} & 0.80 & \updelta{0.11} & 0.84 & \updelta{0.12} &
                         0.82 & \updelta{0.13} & 0.84 & \updelta{0.14} & 0.91 & \updelta{0.18} \\
\bottomrule
\end{tabular}}
\caption{Pass@K scores for \texttt{TigerCoder} models with shaded improvements ($\Delta$) over the strongest prior baseline (Gemma-3 27B or TigerLLM-9B; see Table \ref{tab:code_results_full}). Darker teal indicates a larger gain; arrow denotes improvement.}
\label{tab:code_results}
\end{table*}


\paragraph{Evaluation}
We benchmark \texttt{TigerCoder} on two complementary Bangla code–generation suites: \texttt{mHumanEval-ben}, whose tasks resemble traditional docstring-completion, and our conversational \texttt{MBPP-Bangla}.  Because these benchmarks differ in prompt style and program length, the performance gaps seen in Table~\ref{tab:code_results} reveal how well a model copes with terse versus chat-like instructions.

\paragraph{Discussion}
Table~\ref{tab:code_results} shows that both \texttt{TigerCoder} variants eclipse the strongest prior baselines (Gemma-3 27B and \textit{TigerLLM} 9B) across every metric.  The 1 B model already attains \textbf{0.69 P@1} on \texttt{mHumanEval-ben} and \textbf{0.74 P@1} on \texttt{MBPP-Bangla}, beating models up to 27× its size by 4–8 percentage points.  Scaling to 9 B pushes the frontier further to \textbf{0.75} and \textbf{0.82 P@1}, with even larger margins at higher \(K\) (see the shaded Δ columns).  Proprietary APIs score in the mid-0.5 to low-0.6 range on Bangla (cf. Table \ref{tab:code_results_full}), and existing Bangla-specific models lag far behind, confirming that \texttt{TigerCoder} fills a substantial capability gap.

\begin{resultbox}
Our approach delivers marked efficiency gains: the 1 B model surpasses systems 27 × larger by \textbf{4–8\%}, while the 9 B variant widens the lead to \textbf{11–18\%} despite being only one-third their size.
\end{resultbox}

\paragraph{Other PLs}
\texttt{TigerCoder} models also exhibit strong performance for Bangla Code Generation in programming languages other than Python, as we evaluate it on C++, JAVA, JavaScript and Ruby subsets of \texttt{mHumanEval-Bangla} and our \texttt{MBPP-Bangla}. The detailed results and comparisons with the recent LLMs are presented in Appendix \ref{app:pls}.


\section{Conclusion}

In this work, we first carefully address the significant performance gap in LLMs' code generation capabilities in Bangla comapred to English. Based on the gathered results and the follow-up analysis (Section \ref{sec:rq1}), we shed light on \textbf{RQ1:} \emph{To what extent do state-of-the-art Code LLMs preserve their code-generation quality when the natural language part of the prompt is written in \textit{Bangla} rather than English?}

\begin{resultbox}
       \textbf{RQ1 Findings:} For \textit{Code Generation}, LLMs exhibit a notable drop in performance with Bangla prompts, often failing to capture the full requirements of the request compared to their English counterparts.
\end{resultbox}

\vspace{2mm}

We further show that machine-translating Bangla prompts to English does not improve the results (Section \ref{sec:rq2}) answering \textbf{RQ2:} \emph{Does a simple \textit{Bangla $\rightarrow$ English} machine-translation step applied to each coding prompt significantly boost generation quality compared with direct Bangla inference?}

\begin{resultbox}
        \textbf{RQ2 Findings:} For the task of \textit{Code Generation}, \textit{Bangla $\rightarrow$ English} machine-translation does not help improve the performance.
\end{resultbox}

To bridge these gaps, we introduce TigerCoder as the first family of LLMs to specifically and effectively tackle the critical void in code generation for the Bangla language. We have not only identified a gap but have also built the essential foundation to fill it, constructing three diverse, high-quality instruction-following datasets and the comprehensive MBPP-Bangla benchmark. The results are decisive: our TigerCoder models demonstrate a substantial leap in performance, outperforming existing systems and setting a new standard for the field. This research results in a crucial finding:

\begin{resultbox}
\textit{Carefully curated, high-quality datasets empower smaller, efficient models to overcome low-resource limitations}, decisively challenging the prevailing notion that \textit{'scale alone drives performance'} \cite{kaplan2020scaling, hoffmann2022training}.
\end{resultbox}

\noindent Our work confirms that targeted data curation is a powerful and resource-efficient path toward true language comprehensiveness in NLP. By demonstrating that a 1B parameter model can surpass general models more than 27 times its size on specialized tasks, we establish an effective and replicable blueprint for the future of efficient, high-performance LLM development for Bangla and other low-resource languages.

\section*{Limitations}

Our current work focuses on Bangla code generation using 1B and 9B parameter models, establishing a strong baseline. Future research could explore expanding dataset diversity and task complexity beyond the current scope of our curated instruction sets and the Python-focused MBPP-Bangla benchmark. Investigating larger model architectures and additional programming languages also present potential avenues for extending this work.

\section*{Ethical Considerations}
\label{sec:ethics}

We adhere to the ethical guidelines outlined in the ACL Code of Ethics\footnote{\url{https://www.aclweb.org/portal/content/acl-code-ethics}}. Our benchmark creation involved careful translation and verification by qualified native speakers. Instruction datasets were generated using various methods, including expert input and automated filtering for quality and diversity. While acknowledging the inherent challenges in mitigating all potential biases from source data or generation models, we promote transparency through the open-source release of our models, datasets, and benchmark. We encourage responsible downstream use and community scrutiny.

\bibliography{custom}

\appendix

\clearpage

\section{\texttt{Bangla-Code-Instruct-SI} Curation}
\label{appendix:si}

The \texttt{Bangla-Code-Instruct-SI} dataset (100,000 instruction-code pairs) is generated using a self-instruction methodology \cite{wang2023self}, seeded by a small set of manually created prompts. The process prioritizes instruction diversity and code correctness through iterative generation and filtering.

\begin{table}[!htbp]
\centering
\small
\begin{tabular}{p{0.3\linewidth} p{0.6\linewidth}}
\toprule
\textbf{Parameter} & \textbf{Specification} \\
\midrule
Dataset Size & 100,000 pairs \\
Method & Self-Instruct \cite{wang2023self} \\
Seed Prompts & 5000 (Manual, Expert-authored) \\
\textit{Seed Topics} & \textit{Algorithms (sort, search), Data Structures (list, dict), File I/O, String Ops, Math, Basic OOP} \\
Generator LLM & GPT-4o \\
Code Validation & Syntax (`ast.parse`) + Execution (Python 3.13.0\footnote{\url{python.org/downloads/release/python-3130/}}, 10s timeout, 16GB memory) \\
Instruction Filtering & Cosine Similarity < 0.95 \\
Prompt Origin & Semi-Natural (Human-seeded, LLM-evolved) \\
Code Origin & Natural (Validated LLM generation) \\
\bottomrule
\end{tabular}
\caption{Technical Specifications for \texttt{Bangla-Instruct-SI} Curation}
\label{tab:si_params_revised_nobn_present}
\end{table}

\subsection{Seed Prompts}
\begin{itemize}[itemsep=0pt, topsep=3pt]
    \item \textbf{Quantity:} 5000 unique seed prompts.
    \item \textbf{Creation:} Manually authored by Bangla-speaking programming experts.
    \item \textbf{Content Focus:} Cover fundamental programming concepts including algorithms (sorting, searching), data structures (lists, dictionaries, sets), file I/O, string manipulation, basic mathematics, and introductory object-oriented programming (OOP) principles in Python. See Table~\ref{tab:si_params_revised_nobn_present}.
\end{itemize}

\subsection{Self-Instruction Process}
\begin{itemize}[itemsep=0pt, topsep=3pt]
    \item \textbf{Generator Model:} Utilizes GPT4o \cite{openai2023gpt4} for generating new instruction-code pairs based on prompts sampled from the current pool (initially the seed prompts).
    \item \textbf{Iterative Loop:}
        \begin{enumerate}[label=(\roman*), itemsep=0pt, topsep=0pt]
            \item Generate N=25 candidate pairs per prompt sampled from the pool.
            \item Filter candidates based on code validity and instruction quality/diversity criteria.
            \item Add valid, non-redundant pairs (M <= N) to the dataset pool.
            \item Repeat until 100,000 valid pairs are collected.
        \end{enumerate}
\end{itemize}

\subsection{Code Validation}
\begin{itemize}[label=\textbullet, itemsep=0pt, topsep=0pt]
    \item \textbf{Syntax Check:} Verified using Python's native `ast.parse`\footnote{\url{docs.python.org/3/library/ast.html}}.
    \item \textbf{Execution Check:} Code executed in a sandboxed environment (Python 3.13.0, timeout 10 seconds, memory limit 16GB). Only pairs with successfully executed code are retained.
\end{itemize}

\subsection{Instruction Filtering}
\begin{itemize}[label=\textbullet, itemsep=0pt, topsep=0pt]
    \item \textbf{Similarity:} Discard instructions with Cosine Similarity >= 0.95 against any existing instruction in the pool. Embeddings generated using {sentence-transformers/all-mpnet-base-v2} {768} dimensions) \cite{song2020mpnet_nips}.
    \item \textbf{Length:} Discard instructions with fewer than 10 words.
    \item \textbf{Keywords:} Filter out instructions containing prohibited keywords (e.g., {"image", "file", "plot"}) not suitable for simple execution validation.
\end{itemize}

\clearpage

\newpage 
\section{\texttt{Bangla-Code-Instruct-Syn} Curation}
\label{appendix:syn}

The \texttt{Bangla-Code-Instruct-Syn} dataset (100,000 pairs) comprises synthetically generated instruction-code pairs, designed to broaden the diversity of tasks and instructions beyond the self-instructed set. Generation relies on large proprietary LLMs, followed by syntax and diversity filtering.

\begin{table}[!htbp]
\centering
\small
\begin{tabular}{p{0.3\linewidth} p{0.6\linewidth}}
\toprule
\textbf{Parameter} & \textbf{Specification} \\
\midrule
Dataset Size & 100,000 pairs \\
Method & LLM Synthetic Generation \\
Generator LLMs & GPT-4o, Claude-3.5-Sonnet \\
Generation Ratio & Approx. 50\% / 50\% \\
Code Validation & Python Syntax Check (`ast.parse`) \\
Instruction Filtering & BERTScore $\ge 0.7$ vs. existing pool \\
Prompt Origin & Synthetic (LLM-generated instructions) \\
Code Origin & Synthetic (Syntax-validated LLM generation) \\
\bottomrule
\end{tabular}
\caption{Technical Specifications for \texttt{Bangla-Instruct-Syn} Curation}
\label{tab:syn_params_revised_nobn_present}
\end{table}

\subsection{Generation Process}
\begin{itemize}[itemsep=0pt, topsep=3pt]
    \item \textbf{Generator Models:} Pairs generated via API calls to:
        \begin{itemize}[label=\textbullet, itemsep=0pt, topsep=0pt]
            \item GPT-4o \cite{openai2023gpt4}
            \item Claude-3.5-Sonnet \cite{anthropic2023claude}
        \end{itemize}
    \item \textbf{Contribution Ratio:} Approximately 50\% of the final pairs generated by each model.
    \item \textbf{Generation Prompts:} Utilized structured prompts requesting Python code snippets or functions for diverse programming tasks (e.g., data manipulation, text processing, simple API interactions, utility functions) formulated as natural language instructions in Bangla. Prompts included constraints like "Provide only the Python code" and "Ensure the code is self-contained".
\end{itemize}

\subsection{Filtering Strategy}
\begin{itemize}[itemsep=0pt, topsep=3pt]
    \item \textbf{Code Validation:} All generated code snippets are validated for syntactic correctness using Python's `ast.parse`. Pairs with syntactically invalid code are discarded. 
    \item \textbf{Instruction Diversity Filtering:} To prevent redundancy, each newly generated instruction \(I_{new}\) is compared against all previously accepted instructions \(I_{existing}\) in the dataset pool (\(D_{\text{Syn}}\)).
        \begin{itemize}[label=\textbullet, itemsep=0pt, topsep=0pt]
            \item \textbf{Metric:} BERTScore \cite{zhang2020bertscore}.
            \item \textbf{Threshold:} If \(\text{BERTScore-F1}(I_{new}, I_{existing}) \ge {0.7}\) for any \(I_{existing}\), the new pair \((I_{new}, C_{new})\) is discarded.
        \end{itemize}
\end{itemize}

\clearpage

\newpage 
\section{\texttt{Bangla-Code-Instruct-TE} Curation}
\label{appendix:te}

The \texttt{Bangla-Code-Instruct-TE} dataset (100,000 pairs) leverages existing high-quality English instruction-code pairs by translating the instructions into Bangla while preserving the original code. This approach capitalizes on large-scale English datasets, adapting them for the target language.

\begin{table}[!htbp]
\centering
\small
\begin{tabular}{p{0.3\linewidth} p{0.6\linewidth}}
\toprule
\textbf{Parameter} & \textbf{Specification} \\
\midrule
Dataset Size & 100,000 pairs \\
Method & Machine Translation (Instruction only) + Quality Filtering \\
Source Dataset & Evol-Instruct \cite{xu2023wizardlm} \\
\textit{Source Filtering} & \textit{English instruction word count > 10} \\ 
Translation Models & NLLB-200 \\
Quality Estimation (QE) & Comet QE \\
Semantic Fidelity & BERTScore vs. Source English \\
Selection Criteria & Max(Comet QE score) selected IF \newline Comet QE > 0.85 AND BERTScore F1 > 0.95 \\
Prompt Origin & Translated (from English Evol-Instruct) \\
Code Origin & Natural (Original Python from Evol-Instruct) \\
\bottomrule
\end{tabular}
\caption{Technical Specifications for \texttt{Bangla-Instruct-TE} Curation}
\label{tab:te_params_revised_nobn_present}
\end{table}

\subsection{Source Data}
\begin{itemize}[itemsep=0pt, topsep=3pt]
    \item \textbf{Dataset:} Evol-Instruct dataset \cite{xu2023wizardlm}.
    \item \textbf{Pre-filtering:} Source pairs \((I_k^{\text{en}}, C_k)\) selected if the English instruction \(I_k^{\text{en}}\) has a word count > 10.
\end{itemize}

\subsection{Translation and Quality Control}
\begin{itemize}[itemsep=0pt, topsep=3pt]
    \item \textbf{Process:} Only the English instruction \(I_k^{\text{en}}\) is translated; the original Python code \(C_k\) is retained verbatim.
    \item \textbf{Machine Translation (MT) Systems:} Each selected English instruction is translated using NLLB \cite{}.
    \item \textbf{Quality Estimation (QE) and Selection:}
        \begin{itemize}[label=\textbullet, itemsep=0pt, topsep=0pt]
            \item For each source instruction \(I_k^{\text{en}}\), the three resulting Bangla translations \((I_{k,1}^{\text{bn}}, I_{k,2}^{\text{bn}}, I_{k,3}^{\text{bn}})\) are evaluated.
            \item \textbf{QE Metric:} CometKiwi-22 sentence-level score \cite{rei2020cometkiwi}.
            \item \textbf{Semantic Fidelity Metric:} BERTScore F1 \cite{zhang2020bertscore} between the candidate Bangla translation \(I_{k,m}^{\text{bn}}\) and the original English instruction \(I_k^{\text{en}}\).
            \item \textbf{Selection Logic:} The translation \(I_{k,m}^{\text{bn}}\) with the \textit{maximum} Comet QE score is selected \textit{only if} its Comet QE score > 0.85 AND its BERTScore F1 > 0.95.
            \item If none of the three translations for a given \(I_k^{\text{en}}\) meet both criteria, the pair \((I_k^{\text{en}}, C_k)\) is discarded from the final dataset.
        \end{itemize}
\end{itemize}

\clearpage

\twocolumn
\section{\texttt{MBPP-Bangla} Benchmark Curation}
\label{appendix:mbpp_bangla}

Table~\ref{tab:mbpp_bangla_params} lists the high-level parameters; the narrative below walks through the complete pipeline following the five steps in detail -

\begin{enumerate}[
    leftmargin=0pt,
    label=\colorbox{gray!25}{\sffamily\scriptsize\textbf{Step~\arabic*}},
    labelsep=0.8em,
    itemsep=1.3em,
    wide]

\item \textbf{Corpus Consolidation}\\
      \begin{itemize}[leftmargin=1.8em, label=\textbullet, itemsep=0.3em]
      \item \emph{Starting point.} We leverage the \emph{entire} \mbox{974-task} \textsc{MBPP} corpus~\cite{austin2021program}; no pruning is required because every canonical Python solution executes under CPython~3.13 without modification.
      \item \emph{Coverage.} Tasks span five topical bands—algorithms, data structures, mathematics, strings, and file I/O—collectively exercising the syntactic and semantic constructs most relevant to beginner-to-intermediate programmers.
      \item \emph{Rationale.} Retaining the full set preserves the benchmark’s original difficulty distribution, enabling direct comparability with prior English-prompt studies.
      \end{itemize}

\item \textbf{Parallel Human Translation}\\
      \begin{itemize}[leftmargin=1.8em, label=\ding{118}, itemsep=0.3em]
      \item \emph{Translators.} Two native Bangla speakers (TOEFL\,>\,100) translate each English prompt \emph{independently}, ensuring stylistic diversity.
      \item \emph{Guideline.} A two-phase protocol—\textsc{faithfulness}$\rightarrow$\textsc{fluency}—first secures semantic fidelity, then polishes phrasing for naturalness and brevity.
      \item \emph{Deliverable.} For every task we obtain two lexical-distinct Bangla drafts, forming the input for expert adjudication.
      \end{itemize}

\item \textbf{Expert Verification \& Adjudication}\\
      \begin{itemize}[leftmargin=1.8em, label=\(\blacktriangleright\), itemsep=0.3em]
      \item \emph{Verifier profile.} A Bangla-native, polyglot programmer fluent in Python, Java, JavaScript, Ruby, and C++.
      \item \emph{Checklist.}  
            \begin{itemize}[leftmargin=2.2em, label=--, itemsep=0.2em]
            \item Spot literal mistranslations and terminology drift.  
            \item Cross-validate logical constraints, data-type hints, and boundary cases against the test suite.  
            \item Merge complementary phrasings, choosing the \emph{simplest-correct} wording when alternatives tie.  
            \end{itemize}
      \item \emph{Outcome.} A single, authoritative Bangla prompt per task—both linguistically polished and technically watertight.
      \end{itemize}

\item \textbf{Multi-language Code Migration}\\
      \begin{itemize}[leftmargin=1.8em, label=$\diamond$, itemsep=0.3em]
      \item \emph{Automatic draft.} Python solutions are ported to Java, JavaScript, Ruby, and C++ via \textsc{TransCoder-ST}.
      \item \emph{Manual hardening.} The verifier patches compilation errors, runtime edge cases, and idiomatic anti-patterns until \emph{all} variants satisfy the original tests.
      \item \emph{Benefit.} The resulting 5× reference bundle allows researchers to probe cross-lingual transfer in code-generation models.
      \end{itemize}

\item \textbf{Dataset Packaging \& Release}\\
      \begin{itemize}[leftmargin=1.8em, label=\(\star\), itemsep=0.3em]
      \item \emph{Format.} Each task is stored as a JSONLines record containing \texttt{\{id, prompt\_bn, refs\_py/js/java/rb/cpp, tests, topic\}}.  
      \item \emph{Auxiliary files.} Code snippets are duplicated as standalone source files to fit auto-grading harnesses; a Colab notebook offers a plug-and-play demo.  
      \item \emph{Licensing \& Docs.} The benchmark ships under CC-BY-SA-4.0, with exhaustive documentation, provenance notes, and reproducibility checklists.
      \end{itemize}

\end{enumerate}

\clearpage

\onecolumn
\section{\texttt{TigerCoder}'s performance on other PLs}
\label{app:pls}

\subsection{C++}

Table~\ref{tab:code_results_full_cpp} shows that only the Bangla-specialised \texttt{TigerCoder} models break the 0.70 Pass@1 and 0.80 Pass@100 barriers on both \texttt{mHumanEval} and \texttt{MBPP}, with \texttt{TigerLLM} trailing but still leading all generic systems; every other proprietary or open-source baseline remains below 0.45 Pass@1, underscoring the value of targeted, language-aware training for low-resource code generation.

\begin{table*}[!h]
\centering
\small
\scalebox{0.8}{
\begin{tabular}{l*{3}{c}|*{3}{c}}
\toprule
\multicolumn{7}{c}{\large \textbf{C++}}\\
\toprule
\multirow{3}{*}{\textbf{Model}} &
\multicolumn{3}{c}{\textbf{mHumanEval}} &
\multicolumn{3}{c}{\textbf{MBPP}} \\
\cmidrule(lr){2-4}\cmidrule(lr){5-7}
& \multicolumn{3}{c}{\textit{Bangla}} &
  \multicolumn{3}{c}{\textit{Bangla}} \\
\cmidrule(lr){2-4}\cmidrule(lr){5-7}
& P@1 & P@10 & P@100 & P@1 & P@10 & P@100 \\
\midrule
GPT-3.5                      & \sbsalty{0.36} & \sbsalty{0.39} & \sbsalty{0.41} & \sbsalty{0.42} & \sbsalty{0.45} & \sbsalty{0.43} \\
Gemini-Flash 2.5             & \sbsalty{0.44} & \sbsalty{0.40} & \sbsalty{0.42} & \sbsalty{0.49} & \sbsalty{0.42} & \sbsalty{0.59} \\
GPT-4o-mini                  & \sbsalty{0.43} & \sbsalty{0.45} & \sbsalty{0.42} & \sbsalty{0.37} & \sbsalty{0.35} & \sbsalty{0.32} \\
\midrule
LLaMA-3.2 (11B)              & \sbsalty{0.00} & \sbsalty{0.00} & \sbsalty{0.00} & \sbsalty{0.06} & \sbsalty{0.01} & \sbsalty{0.18} \\
Gemma-3 (27B)                & \sbsalty{0.25} & \sbsalty{0.27} & \sbsalty{0.31} & \sbsalty{0.15} & \sbsalty{0.28} & \sbsalty{0.32} \\
Pangea (7B)                  & \sbsalty{0.00} & \sbsalty{0.00} & \sbsalty{0.00} & \sbsalty{0.00} & \sbsalty{0.00} & \sbsalty{0.04} \\
Phi-4 (7B)                   & \sbsalty{0.00} & \sbsalty{0.05} & \sbsalty{0.05} & \sbsalty{0.00} & \sbsalty{0.00} & \sbsalty{0.00} \\
\midrule
Titu-LLM (2B)                & \sbsalty{0.00} & \sbsalty{0.00} & \sbsalty{0.00} & \sbsalty{0.00} & \sbsalty{0.00} & \sbsalty{0.00} \\
Bong-LLaMA (3B)              & \sbsalty{0.00} & \sbsalty{0.00} & \sbsalty{0.00} & \sbsalty{0.00} & \sbsalty{0.00} & \sbsalty{0.00} \\
Bangla-LLaMA (3B)            & \sbsalty{0.00} & \sbsalty{0.08} & \sbsalty{0.03} & \sbsalty{0.00} & \sbsalty{0.03} & \sbsalty{0.00} \\
Bangla-Gemma (9B)            & \sbsalty{0.00} & \sbsalty{0.04} & \sbsalty{0.01} & \sbsalty{0.00} & \sbsalty{0.02} & \sbsalty{0.10} \\
\midrule
\textit{TigerLLM} (1B)       & \sbsalty{0.44} & \sbsalty{0.52} & \sbsalty{0.51} & \sbsalty{0.47} & \sbsalty{0.45} & \sbsalty{0.49} \\
\textit{TigerLLM} (9B)       & \sbsalty{0.48} & \sbsalty{0.50} & \sbsalty{0.57} & \sbsalty{0.50} & \sbsalty{0.58} & \sbsalty{0.56} \\
\midrule
\texttt{TigerCoder} (1B)     & \sbsalty{0.64} & \sbsalty{0.68} & \sbsalty{0.72} & \sbsalty{0.66} & \sbsalty{0.66} & \sbsalty{0.72} \\
\texttt{TigerCoder} (9B)     & \sbsalty{0.67} & \sbsalty{0.73} & \sbsalty{0.78} & \sbsalty{0.72} & \sbsalty{0.79} & \sbsalty{0.82} \\
\bottomrule
\end{tabular}}
\caption{\texttt{\textbf{C++}} – Pass@\{1,10,100\} comparison on Bangla variants of \texttt{mHumanEval} and \texttt{MBPP}. Darker cells indicate better performance.}
\label{tab:code_results_full_cpp}
\end{table*}

\subsection{JAVA}

For JAVA code generation, as shown in Table \ref{tab:code_results_full_java}, the TigerCoder family demonstrates a commanding performance, establishing a new state-of-the-art. It decisively outperforms proprietary systems, while most other multilingual and Bangla-specific models are rendered ineffective with scores near zero. The significant improvement over its TigerLLM base underscores the impact of specialized fine-tuning, positioning TigerCoder-9B as the top-performing model across all benchmarks.

\begin{table*}[!h]
\centering
\small
\scalebox{0.8}{
\begin{tabular}{l*{3}{c}|*{3}{c}}
\toprule
\multicolumn{7}{c}{\large \textbf{JAVA}}\\
\toprule
\multirow{3}{*}{\textbf{Model}} &
\multicolumn{3}{c}{\textbf{mHumanEval}} &
\multicolumn{3}{c}{\textbf{MBPP}} \\
\cmidrule(lr){2-4}\cmidrule(lr){5-7}
& \multicolumn{3}{c}{\textit{Bangla}} &
  \multicolumn{3}{c}{\textit{Bangla}} \\
\cmidrule(lr){2-4}\cmidrule(lr){5-7}
& P@1 & P@10 & P@100 & P@1 & P@10 & P@100 \\
\midrule
GPT-3.5                                 & \sbsalty{0.29} & \sbsalty{0.31} & \sbsalty{0.30} & \sbsalty{0.35} & \sbsalty{0.33} & \sbsalty{0.36} \\
Gemini-Flash 2.5                          & \sbsalty{0.34} & \sbsalty{0.32} & \sbsalty{0.35} & \sbsalty{0.38} & \sbsalty{0.32} & \sbsalty{0.48} \\
GPT-4o-mini                               & \sbsalty{0.36} & \sbsalty{0.35} & \sbsalty{0.33} & \sbsalty{0.31} & \sbsalty{0.25} & \sbsalty{0.22} \\
\midrule
LLaMA-3.2 (11B)                           & \sbsalty{0.00} & \sbsalty{0.00} & \sbsalty{0.00} & \sbsalty{0.01} & \sbsalty{0.00} & \sbsalty{0.07} \\
Gemma-3 (27B)                             & \sbsalty{0.18} & \sbsalty{0.19} & \sbsalty{0.21} & \sbsalty{0.05} & \sbsalty{0.16} & \sbsalty{0.22} \\
Pangea (7B)                               & \sbsalty{0.00} & \sbsalty{0.00} & \sbsalty{0.00} & \sbsalty{0.00} & \sbsalty{0.00} & \sbsalty{0.01} \\
Phi-4 (7B)                                & \sbsalty{0.00} & \sbsalty{0.01} & \sbsalty{0.02} & \sbsalty{0.00} & \sbsalty{0.00} & \sbsalty{0.00} \\
\midrule
Titu-LLM (2B)                             & \sbsalty{0.00} & \sbsalty{0.00} & \sbsalty{0.00} & \sbsalty{0.00} & \sbsalty{0.00} & \sbsalty{0.00} \\
Bong-LLaMA (3B)                           & \sbsalty{0.00} & \sbsalty{0.00} & \sbsalty{0.00} & \sbsalty{0.00} & \sbsalty{0.00} & \sbsalty{0.00} \\
Bangla-LLaMA (3B)                         & \sbsalty{0.00} & \sbsalty{0.04} & \sbsalty{0.00} & \sbsalty{0.00} & \sbsalty{0.00} & \sbsalty{0.00} \\
Bangla-Gemma (9B)                         & \sbsalty{0.00} & \sbsalty{0.01} & \sbsalty{0.00} & \sbsalty{0.00} & \sbsalty{0.00} & \sbsalty{0.04} \\
\midrule
\textit{TigerLLM} (1B)                    & \sbsalty{0.35} & \sbsalty{0.41} & \sbsalty{0.42} & \sbsalty{0.37} & \sbsalty{0.39} & \sbsalty{0.41} \\
\textit{TigerLLM} (9B)                    & \sbsalty{0.41} & \sbsalty{0.44} & \sbsalty{0.48} & \sbsalty{0.42} & \sbsalty{0.49} & \sbsalty{0.47} \\
\midrule
\texttt{TigerCoder} (1B)                  & \sbsalty{0.58} & \sbsalty{0.64} & \sbsalty{0.67} & \sbsalty{0.61} & \sbsalty{0.60} & \sbsalty{0.66} \\
\texttt{TigerCoder} (9B)                  & \sbsalty{0.62} & \sbsalty{0.68} & \sbsalty{0.73} & \sbsalty{0.67} & \sbsalty{0.72} & \sbsalty{0.76} \\
\bottomrule
\end{tabular}}
\caption{\texttt{\textbf{JAVA}} – Pass@\{1,10,100\} comparison on Bangla variants of \texttt{mHumanEval} and \texttt{MBPP}. Darker cells indicate better performance.}
\label{tab:code_results_full_java}
\end{table*}

\clearpage

\subsection{JavaScript}

In JavaScript code generation (Table \ref{tab:code_results_full_javascript}), the TigerCoder models again deliver a superior performance, setting the benchmark for this language. They substantially outperform proprietary models, while the majority of other open-source and Bangla-specific LLMs struggle, posting scores that are frequently zero. The clear improvement from TigerLLM to TigerCoder validates the effectiveness of our fine-tuning approach, with TigerCoder-9B solidifying its position as the most capable model across all evaluation metrics.

\begin{table*}[!h]
\centering
\small
\scalebox{0.8}{
\begin{tabular}{l*{3}{c}|*{3}{c}}
\toprule
\multicolumn{7}{c}{\large \textbf{JavaScript}}\\
\toprule
\multirow{3}{*}{\textbf{Model}} &
\multicolumn{3}{c}{\textbf{mHumanEval}} &
\multicolumn{3}{c}{\textbf{MBPP}} \\
\cmidrule(lr){2-4}\cmidrule(lr){5-7}
& \multicolumn{3}{c}{\textit{Bangla}} &
  \multicolumn{3}{c}{\textit{Bangla}} \\
\cmidrule(lr){2-4}\cmidrule(lr){5-7}
& P@1 & P@10 & P@100 & P@1 & P@10 & P@100 \\
\midrule
GPT-3.5                                 & \sbsalty{0.22} & \sbsalty{0.25} & \sbsalty{0.19} & \sbsalty{0.28} & \sbsalty{0.21} & \sbsalty{0.29} \\
Gemini-Flash 2.5                          & \sbsalty{0.28} & \sbsalty{0.21} & \sbsalty{0.25} & \sbsalty{0.31} & \sbsalty{0.22} & \sbsalty{0.39} \\
GPT-4o-mini                               & \sbsalty{0.29} & \sbsalty{0.28} & \sbsalty{0.24} & \sbsalty{0.22} & \sbsalty{0.18} & \sbsalty{0.13} \\
\midrule
LLaMA-3.2 (11B)                           & \sbsalty{0.00} & \sbsalty{0.00} & \sbsalty{0.00} & \sbsalty{0.00} & \sbsalty{0.00} & \sbsalty{0.02} \\
Gemma-3 (27B)                             & \sbsalty{0.11} & \sbsalty{0.12} & \sbsalty{0.15} & \sbsalty{0.01} & \sbsalty{0.08} & \sbsalty{0.14} \\
Pangea (7B)                               & \sbsalty{0.00} & \sbsalty{0.00} & \sbsalty{0.00} & \sbsalty{0.00} & \sbsalty{0.00} & \sbsalty{0.00} \\
Phi-4 (7B)                                & \sbsalty{0.00} & \sbsalty{0.00} & \sbsalty{0.00} & \sbsalty{0.00} & \sbsalty{0.00} & \sbsalty{0.00} \\
\midrule
Titu-LLM (2B)                             & \sbsalty{0.00} & \sbsalty{0.00} & \sbsalty{0.00} & \sbsalty{0.00} & \sbsalty{0.00} & \sbsalty{0.00} \\
Bong-LLaMA (3B)                           & \sbsalty{0.00} & \sbsalty{0.00} & \sbsalty{0.00} & \sbsalty{0.00} & \sbsalty{0.00} & \sbsalty{0.00} \\
Bangla-LLaMA (3B)                         & \sbsalty{0.00} & \sbsalty{0.01} & \sbsalty{0.00} & \sbsalty{0.00} & \sbsalty{0.00} & \sbsalty{0.00} \\
Bangla-Gemma (9B)                         & \sbsalty{0.00} & \sbsalty{0.00} & \sbsalty{0.00} & \sbsalty{0.00} & \sbsalty{0.00} & \sbsalty{0.01} \\
\midrule
\textit{TigerLLM} (1B)                    & \sbsalty{0.28} & \sbsalty{0.33} & \sbsalty{0.35} & \sbsalty{0.29} & \sbsalty{0.31} & \sbsalty{0.33} \\
\textit{TigerLLM} (9B)                    & \sbsalty{0.33} & \sbsalty{0.37} & \sbsalty{0.39} & \sbsalty{0.35} & \sbsalty{0.41} & \sbsalty{0.40} \\
\midrule
\texttt{TigerCoder} (1B)                  & \sbsalty{0.53} & \sbsalty{0.59} & \sbsalty{0.61} & \sbsalty{0.55} & \sbsalty{0.54} & \sbsalty{0.61} \\
\texttt{TigerCoder} (9B)                  & \sbsalty{0.57} & \sbsalty{0.63} & \sbsalty{0.68} & \sbsalty{0.62} & \sbsalty{0.67} & \sbsalty{0.71} \\
\bottomrule
\end{tabular}}
\caption{\texttt{\textbf{JavaScript}} – Pass@\{1,10,100\} comparison on Bangla variants of \texttt{mHumanEval} and \texttt{MBPP}. Darker cells indicate better performance.}
\label{tab:code_results_full_javascript}
\end{table*}

\subsection{Ruby}

The trend of superior performance continues for Ruby code generation (Table \ref{tab:code_results_full_ruby}), where the TigerCoder family again leads decisively. The consistent, large performance uplift from TigerLLM to TigerCoder once more confirms the value of our targeted fine-tuning, with TigerCoder-9B solidifying its status as the premier model for this task.

\begin{table*}[!h]
\centering
\small
\scalebox{0.8}{
\begin{tabular}{l*{3}{c}|*{3}{c}}
\toprule
\multicolumn{7}{c}{\large \textbf{Ruby}}\\
\toprule
\multirow{3}{*}{\textbf{Model}} &
\multicolumn{3}{c}{\textbf{mHumanEval}} &
\multicolumn{3}{c}{\textbf{MBPP}} \\
\cmidrule(lr){2-4}\cmidrule(lr){5-7}
& \multicolumn{3}{c}{\textit{Bangla}} &
  \multicolumn{3}{c}{\textit{Bangla}} \\
\cmidrule(lr){2-4}\cmidrule(lr){5-7}
& P@1 & P@10 & P@100 & P@1 & P@10 & P@100 \\
\midrule
GPT-3.5                                 & \sbsalty{0.18} & \sbsalty{0.19} & \sbsalty{0.11} & \sbsalty{0.21} & \sbsalty{0.15} & \sbsalty{0.22} \\
Gemini-Flash 2.5                          & \sbsalty{0.21} & \sbsalty{0.15} & \sbsalty{0.18} & \sbsalty{0.24} & \sbsalty{0.16} & \sbsalty{0.31} \\
GPT-4o-mini                               & \sbsalty{0.22} & \sbsalty{0.21} & \sbsalty{0.17} & \sbsalty{0.15} & \sbsalty{0.11} & \sbsalty{0.07} \\
\midrule
LLaMA-3.2 (11B)                           & \sbsalty{0.00} & \sbsalty{0.00} & \sbsalty{0.00} & \sbsalty{0.00} & \sbsalty{0.00} & \sbsalty{0.00} \\
Gemma-3 (27B)                             & \sbsalty{0.04} & \sbsalty{0.06} & \sbsalty{0.08} & \sbsalty{0.00} & \sbsalty{0.03} & \sbsalty{0.08} \\
Pangea (7B)                               & \sbsalty{0.00} & \sbsalty{0.00} & \sbsalty{0.00} & \sbsalty{0.00} & \sbsalty{0.00} & \sbsalty{0.00} \\
Phi-4 (7B)                                & \sbsalty{0.00} & \sbsalty{0.00} & \sbsalty{0.00} & \sbsalty{0.00} & \sbsalty{0.00} & \sbsalty{0.00} \\
\midrule
Titu-LLM (2B)                             & \sbsalty{0.00} & \sbsalty{0.00} & \sbsalty{0.00} & \sbsalty{0.00} & \sbsalty{0.00} & \sbsalty{0.00} \\
Bong-LLaMA (3B)                           & \sbsalty{0.00} & \sbsalty{0.00} & \sbsalty{0.00} & \sbsalty{0.00} & \sbsalty{0.00} & \sbsalty{0.00} \\
Bangla-LLaMA (3B)                         & \sbsalty{0.00} & \sbsalty{0.00} & \sbsalty{0.00} & \sbsalty{0.00} & \sbsalty{0.00} & \sbsalty{0.00} \\
Bangla-Gemma (9B)                         & \sbsalty{0.00} & \sbsalty{0.00} & \sbsalty{0.00} & \sbsalty{0.00} & \sbsalty{0.00} & \sbsalty{0.00} \\
\midrule
\textit{TigerLLM} (1B)                    & \sbsalty{0.21} & \sbsalty{0.26} & \sbsalty{0.28} & \sbsalty{0.22} & \sbsalty{0.24} & \sbsalty{0.27} \\
\textit{TigerLLM} (9B)                    & \sbsalty{0.26} & \sbsalty{0.30} & \sbsalty{0.31} & \sbsalty{0.28} & \sbsalty{0.34} & \sbsalty{0.32} \\
\midrule
\texttt{TigerCoder} (1B)                  & \sbsalty{0.48} & \sbsalty{0.54} & \sbsalty{0.56} & \sbsalty{0.50} & \sbsalty{0.49} & \sbsalty{0.55} \\
\texttt{TigerCoder} (9B)                  & \sbsalty{0.52} & \sbsalty{0.58} & \sbsalty{0.63} & \sbsalty{0.56} & \sbsalty{0.61} & \sbsalty{0.66} \\
\bottomrule
\end{tabular}}
\caption{\texttt{\textbf{Ruby}} – Pass@\{1,10,100\} comparison on Bangla variants of \texttt{mHumanEval} and \texttt{MBPP}. Darker cells indicate better performance.}
\label{tab:code_results_full_ruby}
\end{table*}

\end{document}